\documentclass[letterpaper, 10 pt, conference]{ieeeconf}  % Comment this line out if you need a4paper

\IEEEoverridecommandlockouts                              % This command is only needed if 
\usepackage{graphics} % for pdf, bitmapped graphics files
\usepackage{epsfig} % for postscript graphics files
\usepackage{amsmath} % assumes amsmath package installed
\usepackage{amssymb}  % assumes amsmath package installed

\title{\LARGE \bf
3D printed cable-driven continuum robots with generally routed cables: modeling and experiments
}

\author{Soumya Kanti Mahapatra$^{1}$, Ashwin K. P.$^{2}$ and Ashitava Ghosal$^{*1}$% <-this % stops a space
\thanks{$^{1}$Department of Mechanical Engineering,
        Indian Institute of Science, Bangalore, India 560012, 
        Email: \tt\{soumyam, asitava\}@iisc.ac.in
        }%
\thanks{$^{2}$ FeatherDyn Pvt. Ltd., Maker village, Kochi, India 683503, Email: \tt ashwin.k.p.2@gmail.com}
\thanks{$^*$ {Corresponding author}}
}

\begin{document}

\maketitle
\thispagestyle{empty}
\pagestyle{empty}

%%%%%%%%%%%%%%%%%%%%%%%%%%%%%%%%%%%%%%%%%%%%%%%%%%%%%%%%%%%%%%%%%%%%%%%%%%%%%%%%
\begin{abstract}

Continuum robots are becoming increasingly popular for applications which require the robots to deform and change shape, while also being compliant. A cable-driven continuum robot is one of the most commonly used type. Typical cable driven continuum robots consist of a flexible backbone with spacer disks attached to the backbone and cables passing through the holes in the spacer disks from the fixed base to a free end. In most such robots, the routing of the cables are straight or a smooth helical curve. In this paper, we analyze the experimental and theoretical deformations of a 3D printed continuum 
robot, for 6 different kinds of cable routings. The results are compared for discrete optimization based kinematic modelling 
as well as static modelling using Cosserat rod theory. It is shown that the experimental results match the theoretical 
results with an error margin of 2\%. It is also shown that the optimization based approach is faster than the one based 
on Cosserat rod theory. We also present a three-fingered gripper prototype where each of the fingers are 3D printed continuum robots with general cable routing. It is demonstrated that the prototype can be used for gripping 
objects and for its manipulation.

%This electronic document is a live template. The various components of your paper [title, text, heads, etc.] are already %defined on the style sheet, as illustrated by the portions given in this document.

\end{abstract}
%%%%%%%%%%%%%%%%%%%%%%%%%%%%%%%%%%%%%%%%%%%%%%%%%%%%%%%%%%%%%%%%%%%%%%%%%%%%%%%%
\section{INTRODUCTION}

\begin{figure}[thpb]
      \centering
      \includegraphics[scale=0.15]{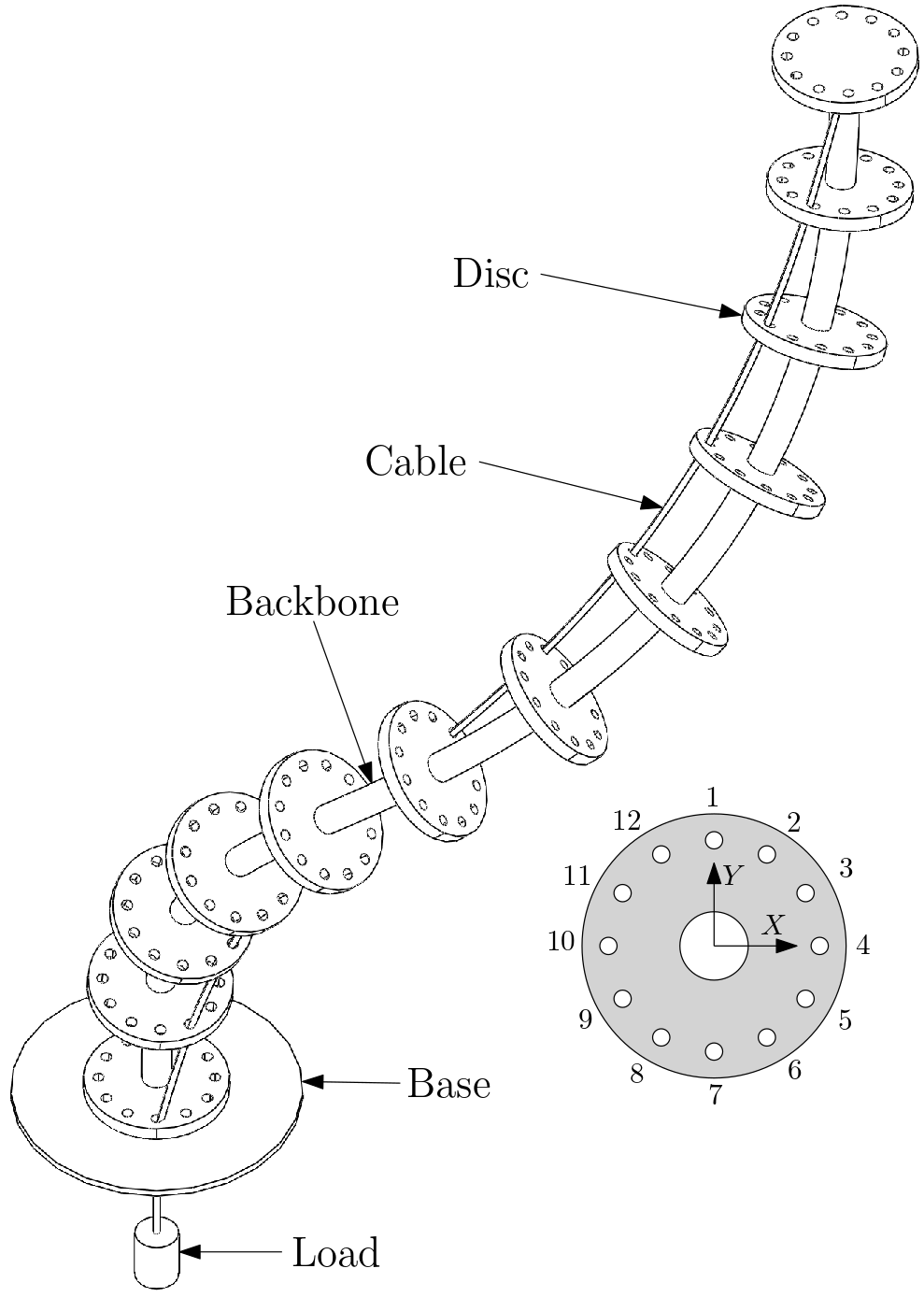}
      \caption{Schematic of a typical cable-driven continuum (CCR) robot}
      \label{CCR}
\end{figure}

Unlike common serial robots with joints and rigid links, a continuum robot uses elastic members and this leads to infinite degrees of freedom. A consequence of the available infinite degrees of freedom is that a continuum robot can theoretically 
take the shape of an arbitrary 3D curve and thus can provide extreme flexibility in attaining various desired poses.
Continuum robots thus finds use in many fields such as in medical devices~\cite{Haghighipanah2015}, remote inspections or in search and rescue in cluttered spaces~\cite{Walker2017} and in space applications~\cite{Tonapi2015}.
Continuum robots have been actuated using cables~\cite{Gravagne2000}, pneumatic tubes~\cite{Trivedi2008}, concentric actuation tubes and rods~\cite{KaiXu2010} and pre-curved concentric tubes~\cite{Rucker2010} to name a few. The actuation enables the continuum robot to achieve a degree of rigidity, which in turn allows the end-effector or tool to be positioned and oriented with sufficient level of accuracy. One of the earliest cable-driven robot was the elephant trunk robot ~\cite{Walker1999},~\cite{Hannan2003} -- a multi-section robot consisting 
of two degree of freedom joints with springs in between and actuated with cables.
In recent developments, cable driven continuum robots are being increasingly studied primarily 
due to their ability to be miniaturized and made light-weight -- some of the well known cable driven robots are 
described in \cite{Tonapi2015}, \cite{Walker1999}, \cite{Renda2012} to name a few.  A cable-driven continuum robot (CCR) consist of a rod like elastic backbone member to which spacer discs are attached at equal intervals as shown in Fig. \ref{CCR}. Holes are present in these discs to route cables through it. These cables start from a base and can end at any of the discs. When the cables are pulled from the bottom, it bends the backbone, and the robot assumes different shapes which depend on the chosen cable routing.

To obtain the shape for a given cable input, various methods have been proposed. The geometric approach shown in [4] 
is the earliest of all and a constant curvature of robot is assumed during the actuation of a straight routed CCR. 
In another approach, the elastic behavior of the backbone~\cite{Yuan2019} have been used to predict the pose of the robot given a particular cable routing. Cosserat rod theory has also been used for this purpose by~\cite{Rucker2011} 
and~\cite{Dehghani2013} and this approach  is claimed to produce very accurate results with very small tip position 
error. This method can also predict pose for general cable routing as shown in~\cite{Starke2017}. In a recent 
optimization~\cite{Ashwin2019} based approach, the backbone is discretised into four-bar linkages and an 
objective function based on the coupler angle is minimized. The main advantage is its geometric nature, although it 
requires an initial guess and the solution may represent a local minima. In this work, we compare the results from 
the two modelling techniques -- the optimization based approach and the Cosserat rod theory based approach. 
The theoretical models are validated against experiments conducted on a 3D printed continuum robot. We 
also present one of the potential applications of the generally routed continuum robot -- a three fingered robotic gripper. 

This paper is organized as follows: in section~\ref{modeling}, we present the optimization based and the Cosserat rod based approaches for arbitrary cable routing. In section~\ref{approach} we present the details of the numerical method used to obtain the pose of the robot for generic cable routing and present the pose and workspace obtained for a few 
generally routed CCRs.  In section~\ref{experiments}, we present the design and fabrication of the continuum robots 
and the results obtained from the prototypes with six general cable routing. A comparison of the experimental and 
numerical results is also shown in section~\ref{experiments}. In section~\ref{prototype}, we present the results 
obtained for a three-fingered gripper constructed with three generally routed CCRs and finally in 
section~\ref{conclusion}, we present the conclusions of this work.  

\section{KINEMATICS OF CCR}
\label{modeling}
In this section, we briefly explain the optimization based method as introduced in~\cite{Ashwin2019} and the Cosserat rod based approach as presented in ~\cite{Rucker2011}.

\subsection{Optimization based approach}
\begin{figure}[thpb]
      \centering
      \includegraphics[scale=0.16]{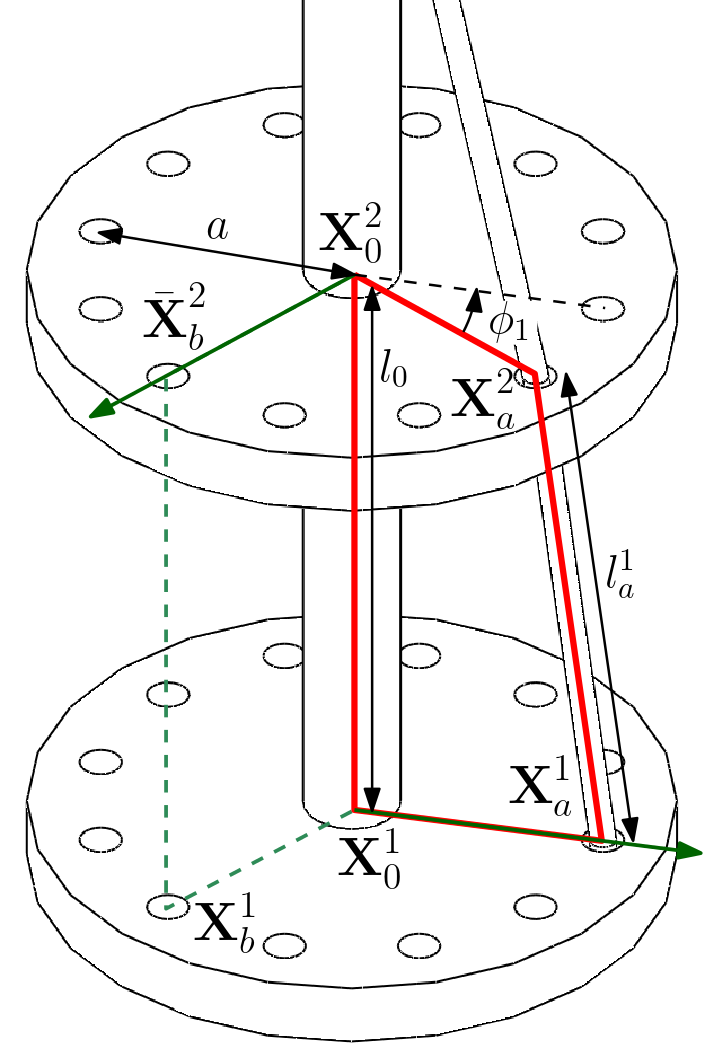}
      \caption{Discretization of generally routed CCR}
      \label{optifig}
\end{figure}

In this approach, a virtual four bar linkage is assumed with vertices $\mathbf{X}_0^1$, $\mathbf{X}_0^2$, $\mathbf{X}_b^2$ and $\mathbf{X}_b^0$ (shown in red colour in Fig. \ref{optifig}),  $\mathbf{X}_0^1 \, \mathbf{X}_a^1$ form 
the base and $\mathbf{X}_0^2 \, \mathbf{X}_a^2$ forms the coupler.
One can obtain the pose of the CCR by minimizing both the coupler angles.
The optimization problem is stated below and this is solved iteratively for all sections starting from the base.

\begin{flalign}
\label{eqnopt}
\nonumber\operatorname*{arg\,min}_{\textbf{x}_0^{i+1},\textbf{x}_a^{i+1}}~\left[\arccos \left( \left(\frac{A}{\Vert A \Vert} \right)\cdot 
\left(\frac{B}{\Vert B \Vert} \right)\right)\right]^2+\\
\left[\arccos \left( \left(\frac{C}{\Vert C \Vert} \right)\cdot 
\left(\frac{D}{\Vert D \Vert} \right)\right)\right]^2
\end{flalign}
\begin{tabular}{c c}
$A = \mathbf{X}_0^{i}-\mathbf{X}_a^{i}$ &$B = \mathbf{x}_0^{i+1}-\mathbf{x}_a^{i+1}$\\
$C = \mathbf{X}_0^{i}-\mathbf{\bar{X}}_b^{i}$ &$D = \mathbf{x}_0^{i+1}-\mathbf{x}_b^{i+1}$
\end{tabular}

Subject to:
\begin{equation}
\label{constraints}
\begin{gathered}
\Vert \textbf{x}_0^{i+1}-\textbf{X}_0^{i} \Vert= l_0\\
\Vert \textbf{x}_a^{i+1}-\textbf{X}_a^{i} \Vert= l_a^i\\
\Vert \textbf{x}_0^{i+1}-\textbf{x}_a^{i+1} \Vert=a\\
\mathbf{\bar{X}}_b^i = a\frac{\left(\mathbf{X}_a^i-\mathbf{X}_0^i \right)\times \left(\mathbf{X}_0^{i+1}-\mathbf{X}_0^i \right)}{\Vert \left(\mathbf{X}_a^i-\mathbf{X}_0^i \right)\times \left(\mathbf{X}_0^{i+1}-\mathbf{X}_0^i \right)\Vert}
\end{gathered}
\end{equation}
Given data: $\mathbf{X}_0^i,\mathbf{X}_0^{i+1},\mathbf{X}_a^i,\mathbf{X}_a^{i+1},l_0,l_a^i,a$

In the above expressions, $\textbf{x}$ represents $\textbf{X}$ after deformation, $l_0$ is the distance between two consecutive discs and  $l_a^i$ is the length of cable in the $i^{th}$ section after deformation, which will be different for each section. In the above,  '$a$' is the distance between the hole and the center
of the disc.

The solution procedure consists of first solving equation (\ref{eqnopt}) for the base linkage.
This provides the co-ordinates of the center of the first disc and the co-ordinates of the hole through which the  
cable is passed, once the cable is given an actuation of $\delta_1$ deformation.
The coupler co-ordinates thus obtained becomes the fixed link of the second segment and the procedure is repeated till the last disk in the CCR. Details of the algorithm can be found in~\cite{Ashwin2019}.  

\subsection{Cosserat rod modeling}
\label{cosserat}
\begin{figure}[thpb]
      \centering
      \includegraphics[scale=0.2]{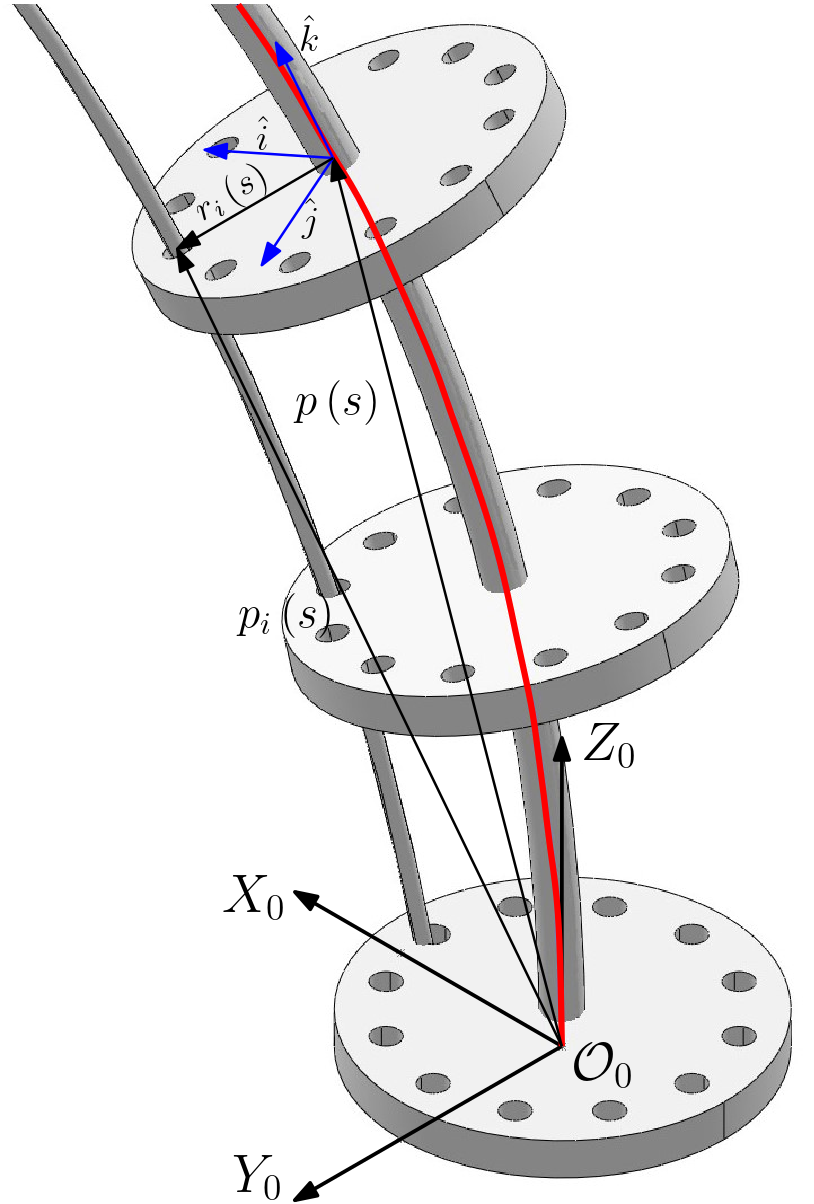}
      \caption{Cosserat rod nomenclature of a CCR}
      \label{cosseratfig}
\end{figure}

In Cosserat rod theory, the elastic backbone is characterized by a continuous 3D curve~\cite{Antman1995}. The cables 
are assumed to be inextensible with no friction at the cable-disc interface and the location of the cable at any 
section remains constant even after actuation. The position of the $i^{th}$ cable at a section in fixed reference coordinate system $\mathcal{O}_0$ is denoted by $\boldsymbol{p}_i\left(s\right)$ and in the local coordinate system is denoted by $\boldsymbol{r}_i\left(s\right)$. The Z-component of $\boldsymbol{r}_i\left(s\right)$ is zero as seen in Fig. \ref{cosseratfig}.

Following the development  in~\cite{Rucker2011}, with no externally applied forces or moments, the set of differential equations that define the position and orientation of the backbone are given by 
\begin{eqnarray}
\label{eqn4}
\dot{\boldsymbol{p}}\left(s\right)&=&\boldsymbol{R}\left(s\right)\boldsymbol{v}\left(s\right)\\
\label{eqn5}
\dot{\boldsymbol{R}}\left(s\right)&=&\boldsymbol{R}\left(s\right)\widehat{\boldsymbol{u}}\left(s\right)
\end{eqnarray}
where $\widehat{\left(\cdot\right)}$ represents  the skew-symmetric matrix made by the elements of $\left(\cdot\right)$, 
$\boldsymbol{p}\left(s\right)$ is any point on the backbone with $s\in\left[0,L\right]$ as the 
arc-length from a fixed reference $\mathcal{O}_0$ with $L$ being the total length of the curve, and 
$\boldsymbol{R}\left(s\right)$ representing the rotation matrix which transforms the local coordinate system 
attached to $\boldsymbol{p}\left(s\right)$ to the fixed $\mathcal{O}_0$.

The rate of change of $\boldsymbol{v}$ and $\boldsymbol{u}$ are given by 
\begin{equation}
\label{eqn9}
\begin{bmatrix}
\dot{\boldsymbol{v}}\\\dot{\boldsymbol{u}}
\end{bmatrix}
=
\begin{bmatrix}
\boldsymbol{K}_{se}+\boldsymbol{A} & \boldsymbol{G}\\
\boldsymbol{G}^{T} & \boldsymbol{K}_{bt}+\boldsymbol{H}
\end{bmatrix}^{-1}
\begin{bmatrix}
\boldsymbol{d}\\\boldsymbol{c}
\end{bmatrix}
\end{equation}

where,

$\boldsymbol{c}=-\widehat{\boldsymbol{u}}\,\boldsymbol{K}_{bt}\,\boldsymbol{u}-\widehat{\boldsymbol{v}}\,\boldsymbol{K}_{se}\left(\boldsymbol{v}-\left[\begin{array}{ccc}
0 & 0 & 1\end{array}\right]^{T}\right)-\boldsymbol{b}$

$\boldsymbol{d}=-\widehat{\boldsymbol{u}}\,\boldsymbol{K}_{se}\left(\boldsymbol{v}-\left[\begin{array}{ccc}
0 & 0 & 1\end{array}\right]^{T}\right)-\boldsymbol{a}$
\begin{equation*}
\begin{tabular}{c c c}
$A_{i}=-\tau_{i}\displaystyle{\frac{\left(\widehat{\dot{\boldsymbol{p}}}_{i}^{b}\right)^{2}}{\left\Vert \dot{\boldsymbol{p}}_{i}^{b}\right\Vert ^{3}}}$,
&$\boldsymbol{A}=\displaystyle{\sum_{i=1}^{n}}A_{i}$,
&$\boldsymbol{G}=-\displaystyle{\sum_{i=1}^{n}}A_{i}\,\widehat{\boldsymbol{r}}_{i}$,
\end{tabular}
\end{equation*}
\begin{equation*}
\begin{tabular}{c c}
$\boldsymbol{H}=-\displaystyle{\sum_{i=1}^{n}}\widehat{\boldsymbol{r}}_i\,A_i\,\widehat{\boldsymbol{r}}_{i}$
&$\boldsymbol{a}=\displaystyle{\sum_{i=1}^{n}}A_{i}\left(\widehat{\boldsymbol{u}}\dot{\boldsymbol{p}}_{i}^{b}+\widehat{\boldsymbol{u}}\dot{r}_{i}+\ddot{\boldsymbol{r}}_{i}\right)$,
\end{tabular}
\end{equation*}
\begin{equation*}
\boldsymbol{b}=\displaystyle{\sum_{i=1}^{n}}\widehat{\boldsymbol{r}}_{i}A_{i}\left(\widehat{\boldsymbol{u}}\dot{\boldsymbol{p}}_{i}^{b}+\widehat{\boldsymbol{u}}\dot{r}_{i}+\ddot{\boldsymbol{r}}_{i}\right)
\end{equation*}
with  $\left(\cdot\right)^{b}$ represents $\left(\cdot\right)$ in body (or local) frame of reference, $n$ is the number of cables, $\tau_i$ is the force applied on the $i^{th}$ cable at the base of the CCR and 
\begin{eqnarray}
\boldsymbol{K}_{se} &=& diag\left( G\,A,  \,G\,A,  \,E\,A\right) \nonumber\\
\boldsymbol{K}_{bt} &= & diag\left(E\,I_{xx}, \,E\,I_{yy},\, G\,I_{zz}\right) \nonumber
\end{eqnarray}
with $E$ and $G$ denote the modulus of elasticity and shear modulus of the material, respectively, $A$ is the cross sectional area and $I_{xx}$, $I_{yy}$ and $I_{zz}$ are the area moments of inertia about the principal axes.

The above differential equations are subjected to boundary conditions and these are obtained as follows:

The CCR is assumed to start from the origin of $\mathcal{O}_0$ and is initially straight before actuation. Hence, 
\begin{eqnarray}
\label{eqn10}
\boldsymbol{p}\left(0\right) = \left[\begin{array}{ccc} 0 & 0 & 0\end{array}\right]^{T}\\
\label{eqn11}
\boldsymbol{R}\left(0\right) = \boldsymbol{I}_3, \; 3\times 3 \; {\rm identity matrix}
\end{eqnarray}
When a cable terminates at $s=L$, it exerts a force and torque on the backbone, tangent to the cable at that point. These  
can be calculated as
\begin{eqnarray}
\label{eqn12}
\boldsymbol{F}_{i}\left(L\right)=-\tau_{i}\displaystyle{\frac{\dot{\boldsymbol{p}_{i}}\left(L\right)}{\left\Vert \dot{\boldsymbol{p}_{i}}\left(L\right)\right\Vert }}\\
\label{eqn13}
\boldsymbol{T}_{i}\left(L\right)=-\tau_{i}\displaystyle{\left(\boldsymbol{R}\left(L\right)\,\boldsymbol{r}_{i}\left(L\right)\right)\widehat{}\,\frac{\dot{\boldsymbol{p}_{i}}\left(L\right)}{\left\Vert \dot{\boldsymbol{p}_{i}}\left(L\right)\right\Vert }}
\end{eqnarray}

Hence, $\boldsymbol{u}\left(L\right)$ and $\boldsymbol{v}\left(L\right)$ should be such that
\begin{equation}
\label{eqn14}
\boldsymbol{n}\left(s\right) = \boldsymbol{F}\left(s\right) \;\&\; \boldsymbol{m}\left(L\right) = \boldsymbol{T}\left(L\right)
\end{equation}
where $\boldsymbol{n}\left(L\right)$, $\boldsymbol{m}\left(L\right)$ represents the internal force and moment applied on the backbone per unit length.

One can solve the first order differential equations (\ref{eqn4}), (\ref{eqn5}) and (\ref{eqn9}) with the boundary conditions (\ref{eqn10}) - (\ref{eqn13}) to obtain the pose of the CCR after actuation. Details of this approach can be found in~\cite{Rucker2011}.

\section{ALGORITHM AND NUMERICAL APPROACH}
\label{approach}

\begin{figure}[thpb]
      \centering
      \includegraphics[scale=0.2]{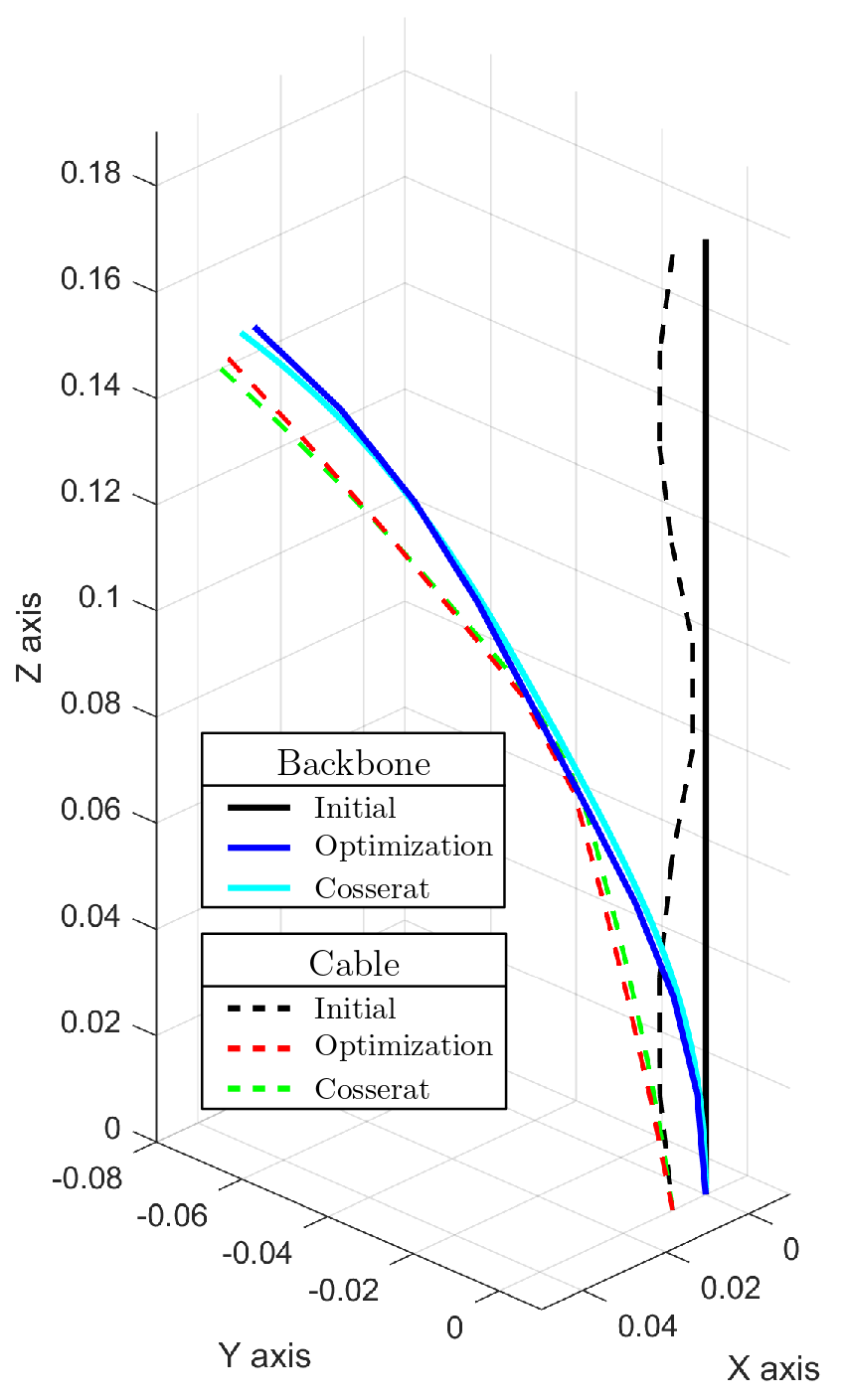}
      \caption{Comparison of simulation results}
      \label{simres}
\end{figure}

The optimization problem (\ref{eqnopt}) was solved using \texttt{fmincon} function in MATLAB\textsuperscript \textregistered \cite{MATLAB:2018a}. For the Cosserat rod model, differential equations (\ref{eqn4}), (\ref{eqn5}) and (\ref{eqn9}) are solved in MATLAB\textsuperscript \textregistered  using shooting method~\cite{John2019}.
For numerical simulation as well as the validation experiments, six cable routings (I - VI) were chosen which are provided in 
Table I -- the location of the cable on the disc follow the numbering same as in Fig. \ref{CCR} with 
Disc 1 as the base of the CCR and Disc 10 as the tip. Routing I is straight, II is helical and the rest are arbitrary paths.
A sample numerical results for routing VI is plotted in Fig. \ref{simres}. A load of 400 g is applied on the cable (input for Cosserat model) and the change in length of the cable was 5.2\% (input for optimization model). The initial configuration is shown in black. The blue and cyan  solid lines are the solutions for the backbone from optimization and the Cosserat 
rod model respectively. The initial and the final cable configurations are also shown in broken lines in Fig. \ref{simres}.  The solution 
to the optimization problem and differential equations on an average takes about 2.5 sec and  10.5 sec, respectively, to complete on an Intel processor at 3.1 GHz and 16 GB RAM.

\begin{table}[h]
\label{table1}
\caption{Cable positions of various routings}
\begin{center}
\begin{tabular}{|c|c|c|c|c|c|c|c|c|c|c|}
\hline 
 & \multicolumn{10}{c|}{Cable position}\tabularnewline
\hline 
Disc No.$\rightarrow$& \textbf{1} & \textbf{2} & \textbf{3} & \textbf{4} & \textbf{5} & \textbf{6} & \textbf{7} & \textbf{8} & \textbf{9} & \textbf{10}\tabularnewline
\hline
I (Straight) & 1 & 1 & 1 & 1 & 1 & 1 & 1 & 1 & 1 & 1\tabularnewline
\hline 
II (Helical) & 4 & 3 & 2 & 1 & 12 & 11 & 10 & 9 & 8 & 7\tabularnewline
\hline 
III (General) & 4 & 3 & 2 & 1 & 1 & 1 & 1 & 2 & 3 & 4\tabularnewline
\hline 
IV (General) & 10 & 10 & 10 & 10 & 10 & 10 & 9 & 8 & 7 & 6\tabularnewline
\hline 
V (General) & 4 & 3 & 2 & 1 & 12 & 11 & 10 & 9 & 9 & 9\tabularnewline
\hline 
VI (General) & 4 & 5 & 6 & 7 & 8 & 8 & 7 & 6 & 5 & 4\tabularnewline
\hline 
\end{tabular}
\end{center}
\end{table}

%For the optimization we can get solutions with fewer sections to capture the pose as compare to Cosserat rod model. This also leads to faster solutions. The optimization is material independent, and hence it can applied for CCR whose material properties or applied loads are unknown as displacements are easier to measure.

\begin{figure}[thpb]
      \centering
      \includegraphics[scale=0.16]{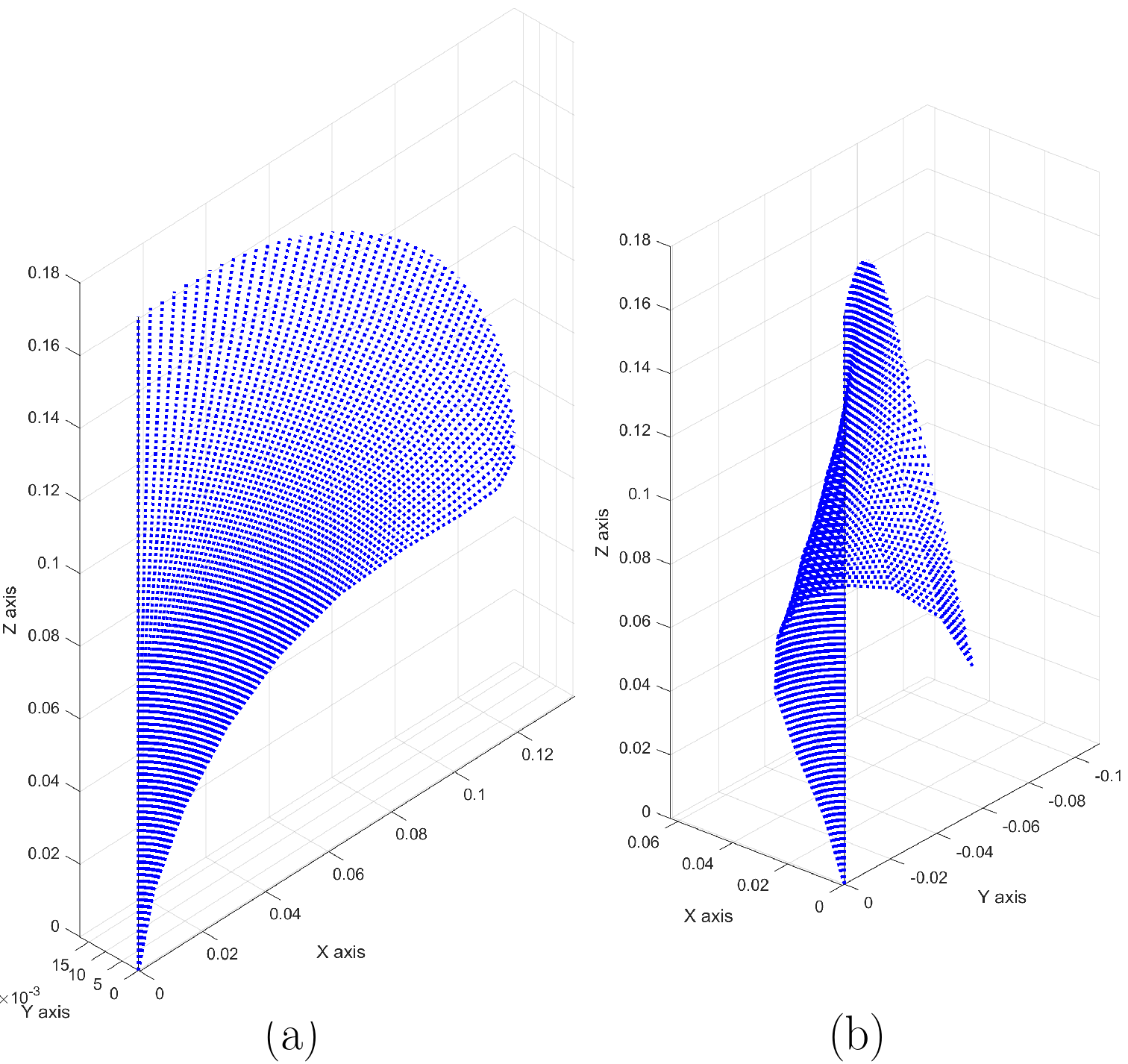}
      \caption{Workspace for: (a) Routing IV, (b) Routing VI}
      \label{workspace}
\end{figure}

The optimization based numerical approach was used to obtain the workspace of two CCRs -- the cable routings IV and VI, given in Table I -- and these are shown in Fig. \ref{workspace}. The workspaces were obtained in 38 seconds
and it can be seen that interesting workspaces can be obtained using general cable routing.

\section{EXPERIMENTAL VALIDATION}
\label{experiments}
\begin{figure*}[thpb]
      \centering
      \includegraphics[scale=0.78]{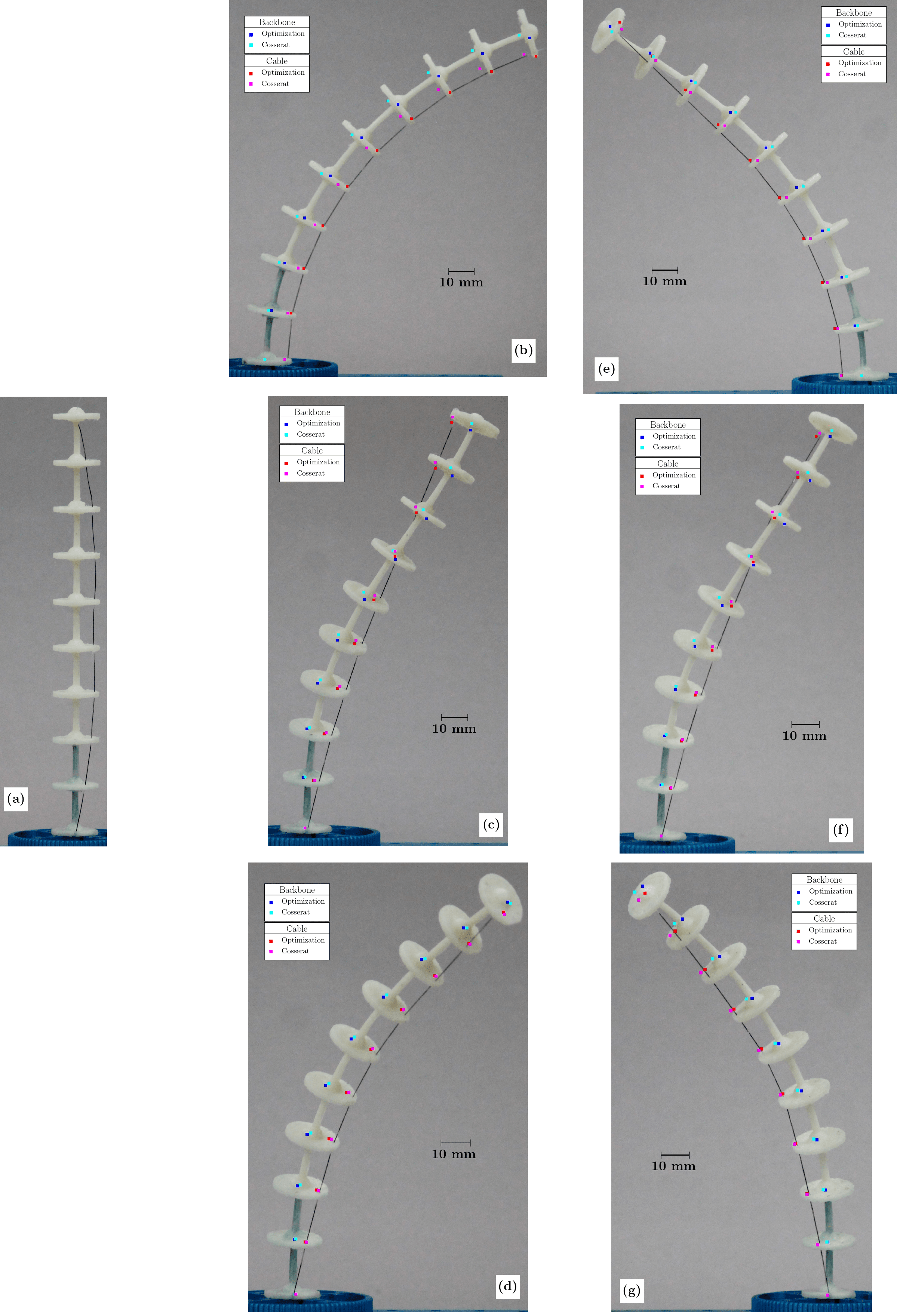}
      \caption{(a) Prototype 3D printed CCR. Comparison for  (b) Routing I, (c) Routing II, (d) Routing III, (e) Routing IV, (f) Routing V, (g) Routing VI}
      \label{result}
\end{figure*}

For the purpose of validation of the results obtained numerically a prototype 3D printed continuum robot was 
fabricated using ABS material (see Fig. \ref{result} (a)). The backbone of the CCR 
has a length,  $L$, of 180 mm and with diameter, $d$, of 3 mm. There are a total of 10 circular discs, 2 mm thick with 12 equally spaced holes at a distance of 8 mm from the center. For ABS, the modulus of elasticity, $E$, used is 1.1 GPa,  Poisson's ratio, $\nu$, is found to be 0.3~\cite{Lay2019}. Very thin fishing wires were used as cables and the 
arrangement is similar to Fig. \ref{CCR}. A weight of 400 g was attached to the cable at the base and the corresponding change in length of the total wire inside the CCR was also measured and is presented in Table II. To validate the numerical results, images of the CCR prototype are taken after actuation and the simulated results are super-imposed onto the image at each disc as shown in Fig. \ref{result} -- the red, blue dots are for optimization results, cyan and magenta dots show the results for Cosserat rod theory and the actual CCR is as shown. The center of the disc at the base is chosen as $\mathcal{O}_0$ with Y-axis towards the right of the figure and Z-axis pointing upwards.
\begin{table}
\label{table2}
\caption{Experimental details}
\begin{center}
\begin{tabular}{|p{1.7cm}|c|c|c|c|c|c|}
\hline  
Routing & I & II & III & IV & V & VI\tabularnewline
\hline 
Load & 400 g & 400 g & 400 g & 400 g & 400 g & 400 g\tabularnewline
\hline 
\% reduction of cable length & 5.5 & 7 & 5.7 & 5.3 & 7 & 5.2\tabularnewline
\hline 
Max error (Optimization) & 0.91 & 1.71 & 1.65 & 1.12 & 1.99 & 1.45\tabularnewline
\hline 
Max error (Cosserat rod) & 1.89 & 1.59 & 1.81 & 1.33 & 1.26 & 1.87\tabularnewline
\hline 
\end{tabular}
\end{center}
\end{table}
%\begin{figure}[thpb]
%      \centering
%      \includegraphics[scale=0.25]{Straight.png}
%      \caption{Result for cable routing I (Straight)}
%      \label{straight}
%\end{figure}
%\begin{figure}[thpb]
%      \centering
%      \includegraphics[scale=0.23]{Helical 30.png}
%      \caption{Result for cable routing II (Helical)}
%      \label{helical}
%\end{figure}
%
%\begin{figure}[thpb]
%      \centering
%      \includegraphics[scale=0.21]{30x3_0x3_-30x3.png}
%      \caption{Result for cable routing III (General I)}
%      \label{strange}
%\end{figure}
%\begin{figure}[thpb]
%      \centering
%      \includegraphics[scale=0.23]{0x5_30x4.png}
%      \caption{Result for cable routing IV (General II)}
%      \label{strange2}
%\end{figure}
%\begin{figure}[thpb]
%      \centering
%      \includegraphics[scale=0.23]{30x7_0x2.png}
%      \caption{Result for cable routing V (General III)}
%      \label{strange3}
%\end{figure}
%\begin{figure}[thpb]
%      \centering
%      \includegraphics[scale=0.23]{-30x4_0x1_30x4.png}
%      \caption{Result for cable routing VI (General IV)}
%      \label{strange4}
%\end{figure}

Figure \ref{result} (b) - (g) shows the comparison of the optimization problem and Cosserat rod simulations with the experimental results. The maximum errors for the backbone as compared to actual prototype for each method are stated in Table II. The maximum error along the backbone is less than 3.6 mm for both cases, i.e., an error of less than 2\% of the total length of the CCR. The simulation results can be seen to match the experimental results for the 3D printed CCR very closely.

\section{A THREE-FINGERED CCR GRIPPER}
\label{prototype}
\begin{figure}[thpb]
      \centering
      \includegraphics[scale=0.13]{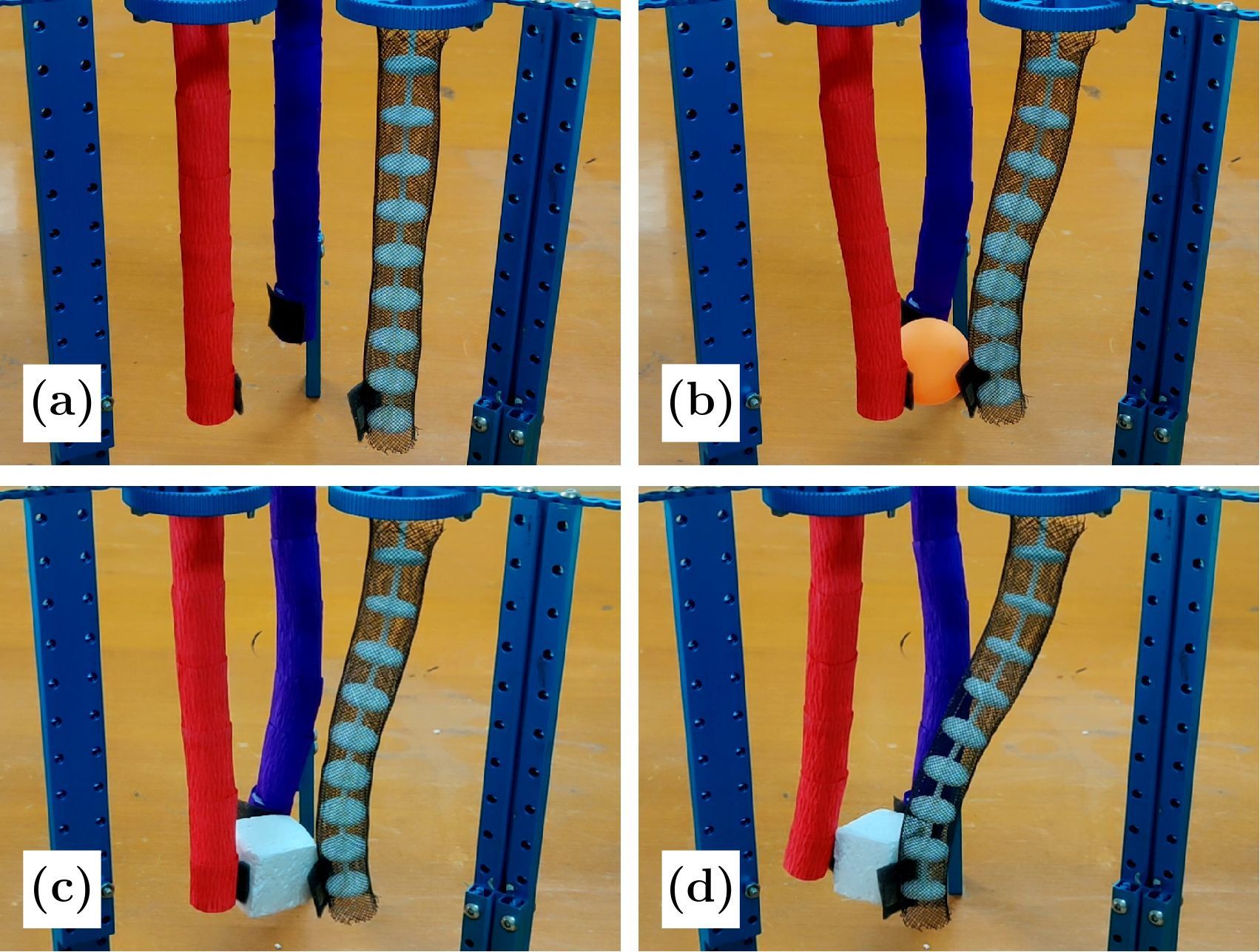}
      \caption{Three fingered gripper: (a) Initial configuration, (b) Gripping a spherical object, (c) Gripping a cube, (d) Manipulating a cube -- screenshots from video.}
      \label{finger}
\end{figure}

A 3-finger gripper was prepared with three CCR attached to a frame and kept inverted as seen in Fig. \ref{finger} (a). Thin foam pads were attached to the tip of all CCR. Using the gripper, small spherical and cubical objects were gripped and manipulated as shown 
as shown in figures \ref{finger}(b)-(d). The attached video shows the three-fingered 
gripper gripping a sphere and a cube. The manipulation of the cube is also shown in the video.

Due to the general nature of the cable routing, we can generate unconventional shapes for CCR which can be used to manipulate objects of various shapes. Also, since the CCR is compliant, such a gripper can be employed to handle delicate objects. Experiments with arbitrary shaped objects are continuing.

\section{CONCLUSIONS}
\label{conclusion}
This paper deals with obtaining the pose of a cable-driven continuum robot with general cable routing. An optimization based approach based on discretizing the continuum robot with four-bar mechanisms and minimizing the coupler 
angle is presented. The results from the optimization based approach are compared with Cosserat rod theory based
approach and it is shown that the results are very similar. The optimization based approach is purely kinematic analysis 
while the Cosserat rod model involves static analysis. The main advantage of the optimization based approach is 
its geometric nature and does not use material properties. The main disadvantage is the requirement of initial guess and 
local minimum nature of the solutions. The solution to static analysis requires slightly more involved computation and 
it captures the physics of the mechanism better by considering the material properties.

To validate the numerical results a 3D printed CCR was fabricated and 6 different cable routings were explored. 
From the results, it can be seen that the simulations are in good agreement with the experiments with error less than 
2\% of the total length of the CCR. Some of the sources of error are anisotropy of the prototype due to directional 
nature of 3D printing resulting in error in the chosen values of the material properties used in the numerical simulations, 
small amounts of friction in the disc-cable interface, and the orientation of the image capture device used to obtained the 
location of the backbone. A three-fingered gripper was constructed with generally routed CCRs. It is shown that the 
three-fingered gripper can grip and manipulate objects. 

In conclusion, we believe a combination of the optimization based and the Cosserat rod theory based approach 
is ideal to estimate the pose of cable driven continuum robots and future work is aimed towards developing such a unified
approach. We are also extending this work to other applications where generally routed CCRs are more advantageous.

\addtolength{\textheight}{-12cm}   % This command serves to balance the column lengths
\section*{ACKNOWLEDGMENT}

We would like to thank UTSAAH Lab and Design Innovation Center (DIC) at CPDM, IISc Bangalore for fabricating the 3D printed prototypes for experimentation.
%%%%%%%%%%%%%%%%%%%%%%%%%%%%%%%%%%%%%%%%%%%%%%%%%%%%%%%%%%%%%%%%%%%%%%%%%%%%%%%%

\end{document}